\documentclass[a4paper]{article}

\usepackage{INTERSPEECH2021}

\title{Polyphone Disambiguation in Mandarin Chinese with Semi-Supervised Learning}
\name{Yi Shi$^*$\thanks{$^*$ Equal Contribution}, Congyi Wang$^{*\dagger}$\thanks{$^\dagger$  Corresponding Author}, Yu Chen, Bin Wang}
\address{Xmov, China}
\email{\{shiyi,wangcongyi,yuchen,wangbin\}@xmov.ai}

\begin{document}

\maketitle
\begin{abstract}
The majority of Chinese characters are monophonic, while a special group of characters, called polyphonic characters, have multiple pronunciations. As a prerequisite of performing speech-related generative tasks, the correct pronunciation must be identified among several candidates. This process is called Polyphone Disambiguation. Although the problem has been well explored with both knowledge-based and learning-based approaches, it remains challenging due to the lack of publicly available labeled datasets and the irregular nature of polyphone in Mandarin Chinese. In this paper, we propose a novel semi-supervised learning (SSL) framework for Mandarin Chinese polyphone disambiguation that can potentially leverage unlimited unlabeled text data. We explore the effect of various proxy labeling strategies including entropy-thresholding and lexicon-based labeling. Qualitative and quantitative experiments demonstrate that our method achieves state-of-the-art performance. In addition, we publish a novel dataset specifically for the polyphone disambiguation task to promote further research.
\end{abstract}

\section{Introduction}
In a Mandarin Chinese TTS system, a key step of Grapheme-to-Phoneme conversion is to identify the correct pronunciation of each character. The major obstacle in solving the problem is posed by polyphone ambiguity, i.e. some characters have distinctive pronunciations under different contexts. 

\subsection{Rule-based approaches}
\begin{figure}
    \centering
    \includegraphics[width=8.5cm]{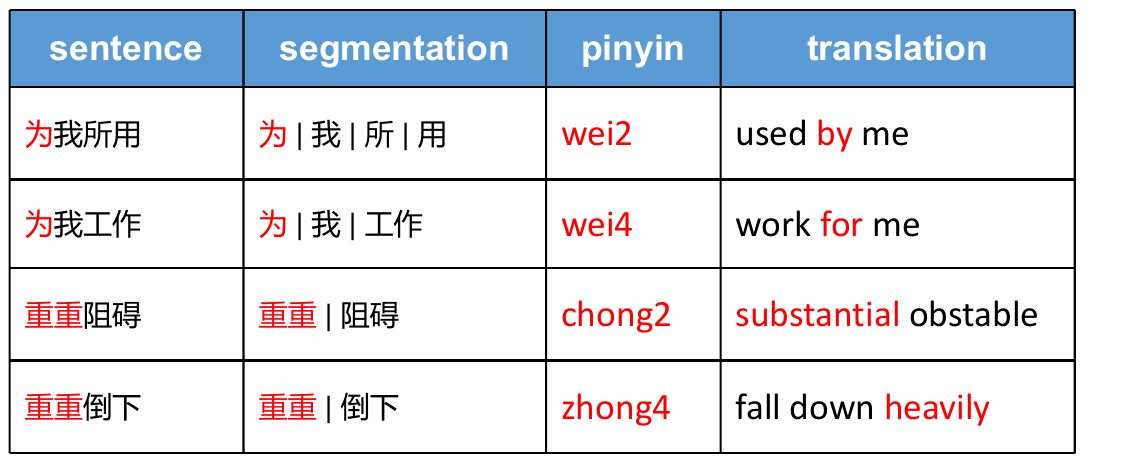}
    \caption{A classic case of pronunciation ambiguity caused by a polyphonic character. The above examples are selected from our evaluation set.}
    \label{fig:f1}
\end{figure}

Early studies \cite{old} focus on building knowledge-based pipelines in which the pronunciations are determined by a sequence of pre-designed rules. Such measures typically segment sentences into words and eliminate any potential pronunciation ambiguity on words that can be matched in a dictionary, then an expert system is utilized in solving the remaining polyphones. Despite the low computational cost, extra information such as POS tag is usually mandatory for high-quality rule-based analysis. Furthermore, knowledge-based approaches are prone to fail on polyphonic characters when they are part of words with distinctive meanings, as shown in Figure \ref{fig:f1}. 

\subsection{Learning-based approaches}
Deep Neural Networks have risen to be the default method to solve problems in natural language and speech processing \cite{oord2016wavenet,trans}. 
Shan et al. \cite{blstm} leverage extra POS tags and implement a bi-directional LSTM layer to model neighboring contexts of the target character. In \cite{distant}, a convolutional seq2seq model proposed by \cite{convseq2seq} is trained on audio-generated texts contained polyphonic characters. Cai et al. \cite{MLEF} treat the Polyphone Disambiguation as a classification problem and generate combined character embedding using a bi-directional LSTM layer and Tencent Chinese Word2Vec embeddings\cite{song2018}. 

In recent years, the introduction of BERT \cite{bert} and similar unsupervised pre-training methods \cite{roberta,albert} take advantage of rich unlabeled text data and train a neural network on performing pre-text tasks such as cloze tests. 
Then the pre-trained model can be connected with simpler neural nets and get fine-tuned on downstream tasks. Dai et al. \cite{polybert} are the first to apply a pre-trained Chinese Bert on the polyphone disambiguation problem. 


These advancements are mainly contributed by the application of supervised learning on increasingly larger models and additional commercial datasets. Nonetheless, very few datasets \cite{korean} for polyphone disambiguation are made available to the public. The existing labeled dataset is also relatively small and unbalanced. In short, the lack of data constitutes an obstacle to the development of Polyphone Disambiguation.


\subsection{Semi-PPL: Semi-supervised Learning for Polyphone Disambiguation with Pseudo Labels and Consistency Loss}
In this paper, we propose \textbf{P}olyphone \textbf{P}seudo \textbf{L}abels (Semi-PPL), a \textbf{Semi}-supervised Learning approach for task of Polyphone Disambiguation. The proposed system consists of two parts: the prediction network built upon a pre-trained language model and the SSL mechanism that is designed to assign pseudo labels. Unlike the previous study \cite{polybert} that directly use a computationally expensive Bert, we adopt a tiny-Electra \cite{electra} to compute character context representations. 

In Table \ref{tab:1}, the base prediction model itself has demonstrated performance that surpasses all recent learning-based methods \cite{bert,MLEF,korean} on our evaluation set. Furthermore, we are the first to apply Semi-supervised Learning (SSL) on the Polyphone Disambiguation task to reduce the dependence on expensive manually labeled data. Pseudo-labeling \cite{fixmatch,meanteacher,chen2020big,chen2020simple,chen2020exploring,berthelot2019mixmatch,berthelot2019remixmatch,bach,he2019moco,chen2020mocov2} is among the most successful SSL strategies in solving computer vision problems. Pseudo-labeling typically starts by acquiring two models \cite{meanteacher}. One model serves as the teacher, assigning labels to unlabeled data that satisfy criteria and expand the labeled sets. The other one takes the role of the student and learns from the labeled training set combined with pseudo labeled data. We adapt pseudo-labeling to the current task by designing an entropy criterion for label assignment to polyphonic character embedding. Pseudo labels are only assigned to unlabeled texts when the base model is 'certain', which is measured by information entropy of output probability distribution. We also propose a language consistency regularization mechanism that imposes output similarity between augmented versions of texts contained polyphonic characters. We publish a dataset that includes 300K labeled texts and an evaluation set with 1187 difficult sentences that are manually annotated.


In summary, the contribution of our work is fourfold: \begin{enumerate} \item We propose an end-to-end Deep Neural Network structure that achieves state-of-the-art performance on polyphone disambiguation with a light weighted model. Our model greatly relieves the hardware requirement compared to other Bert based approach, making it possible to be deployed in the large-scale scenario. \item We introduce Semi-PPL, a Semi-Supervised learning paradigm to tackle the challenging Polyphone Disambiguation problem, boosting the performance significantly only using cheap unlabeled data. \item We design a novel consistency regularization mechanism specifically for character-level tasks on Chinese texts. As shown in the experiments, it greatly enhances the accuracy of the model on difficult short sentences. \item We publish a sizeable dataset for Mandarin Chinese polyphone disambiguation and provide a benchmark with recent learning-based approaches.  \end{enumerate}

\section{Method}
The front end of a Mandarin Chinese Text-to-Speech (TTS) system often comprises a text segmenter and a Grapheme-to-Phoneme (G2P) module.  
It is a common practice in the industry to leverage an existing text segmenter and a Chinese word-to-pinyin dictionary. The raw texts are first segmented into words, then the pinyin of most characters are determined after checking the dictionary. As mentioned in \cite{MLEF}, the pronunciation ambiguation that exists in monophonic character words (MCW) is often eliminated in this step, except only a few exceptions. As is depicted in Figure \ref{fig:f1}, certain 'difficult' cases are unsolvable by traditional method. In such cases, polyphonic characters are usually part of words look the same but have multiple meanings and pronunciations. 

In this paper, Polyphone Disambiguation is considered as a character-wise classification task in which all of the target polyphonic characters share a classifier. Each possible pronunciation is viewed as a single class.

\subsection{Base Model Architecture}
\begin{figure}
    \centering
    \includegraphics[width=8.8cm]{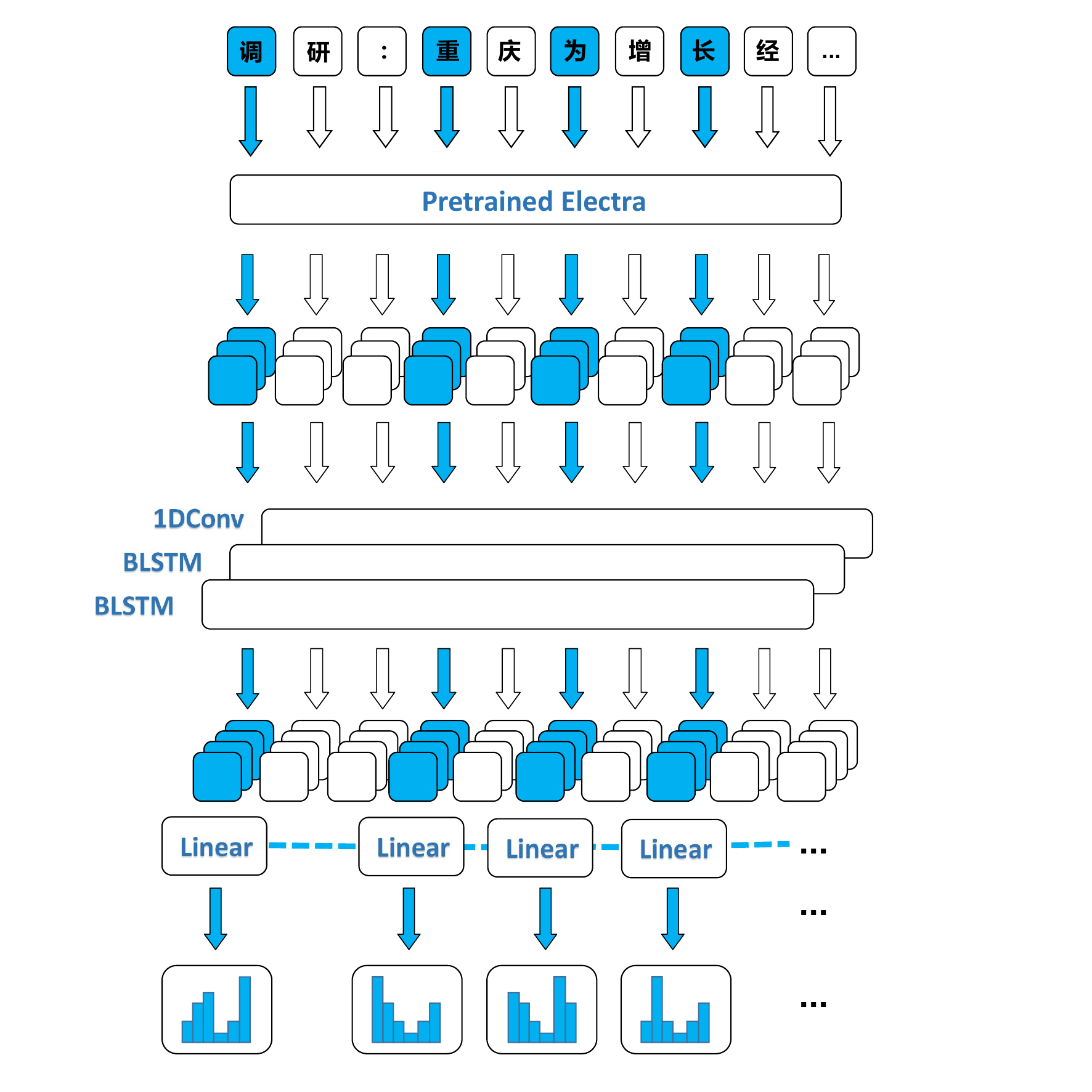}
    \caption{The network architecture of our proposed approach. The characters in the blue boxes are polyphonic characters. }
    \label{fig:f2}
\end{figure}

The model of our approach can be divided into two parts: a pre-trained Electra model and the prediction network as depicted in Figure \ref{fig:f2}. The pre-trained Electra transforms a sequence of Chinese characters into character-level embeddings incorporated with contextual information. The Convolution-BLSTM prediction net serves the purpose of further learning sentence-level knowledge. The output hidden vectors of the selected polyphonic characters are accepted as the input of a series of dense layers to produce the final probability distribution. 

\subsubsection{Electra}
Recently, connecting the existed networks with a pre-trained Bert is the de facto way of improving performance in the field of NLP. The input taken by a Bert during its unsupervised training stage consists of a sequence of tokens $[x_0,x_1,x_2,x_3,...x_n]$ with each token having a probability to be replaced by a [MASK] placeholder. Then a transformer model learns to solve a cloze test, i.e predict the identity of the original token. However, part of Bert's success is due to its heavyweight structure, making the computational cost too high from the perspective of industrial deployment. The authors of \cite{electra} introduce a more efficient pre-training task where the model learns to differentiate real sentence tokens apart from synthetic replacements. The key innovation lies in two aspects:  \begin{enumerate}
    \item It improves the training efficiency given all tokens in an input string, instead of randomly masked ones, can be used during the unsupervised training phase.
    \item The input mismatch is greatly alleviated. More specifically, [MASK] tokens are only seen when Bert is pre-trained on an unlabeled text corpus. Electra, on the other hand, accepts a noisy version of real text data.
\end{enumerate}  
The modifications made by the Electra model result in a reduction in weight size and an improvement in the performance of downstream tasks. By adopting a tiny-Electra pre-trained on large Chinese Corpus \cite{cui-etal-2020-revisiting}, our base model has a significantly smaller weight size and achieves competitive performance compared to methods utilizing Bert \cite{polybert} as demonstrated in the experiments section.

\subsubsection{Convolution BLSTM based prediction network}
Previous research\cite{polybert} suggests that using two sequential BLSTM layers after a pre-trained Bert further improves task performance. This setting even beats the popular transformers. We speculate that the phenomenon is contributed mainly by the difference behind self-attention and contextual information propagation realized by hidden latent. The usage of two different mechanisms covers sentence-level semantic knowledge more thoroughly. Inspired by DeepSpeech2 \cite{DeepSpeech2}, we connect two sequential BLSTM layers with a 1D-convolutional layer and two fully-connected linear layers. LayerNorm \cite{ba2016layer} are applied before each linear layer. We use GeLU \cite{hendrycks2016gaussian} as the activation function. The weight of linear layers is shared among all candidate polyphonic characters. Moreover, We ignore all impossible classes (the pronunciations that don't belong to the target character) by setting the scores of the corresponding classes as $\mathbf{-inf}$ before the final softmax layer.

\subsection{Semi-Supervised Learning}
Multiple studies \cite{fixmatch,berthelot2019mixmatch,berthelot2019remixmatch,xie2019unsupervised} in the field of Computer Vision have proved the effectiveness of Semi-Supervised Learning (SSL). SSL is expected to enhance task performance by leveraging numerous unlabeled data while only a limited quantity of labeled data are available. The introduction of our proposed method is divided into two subsections. Section \ref{cl} explains the details of the consistency loss applied on input text data contained polyphonic characters, while Section \ref{pl} covers several viable options for pseudo label assignment. 

\begin{figure}
    \centering
    \includegraphics[width=8.1cm]{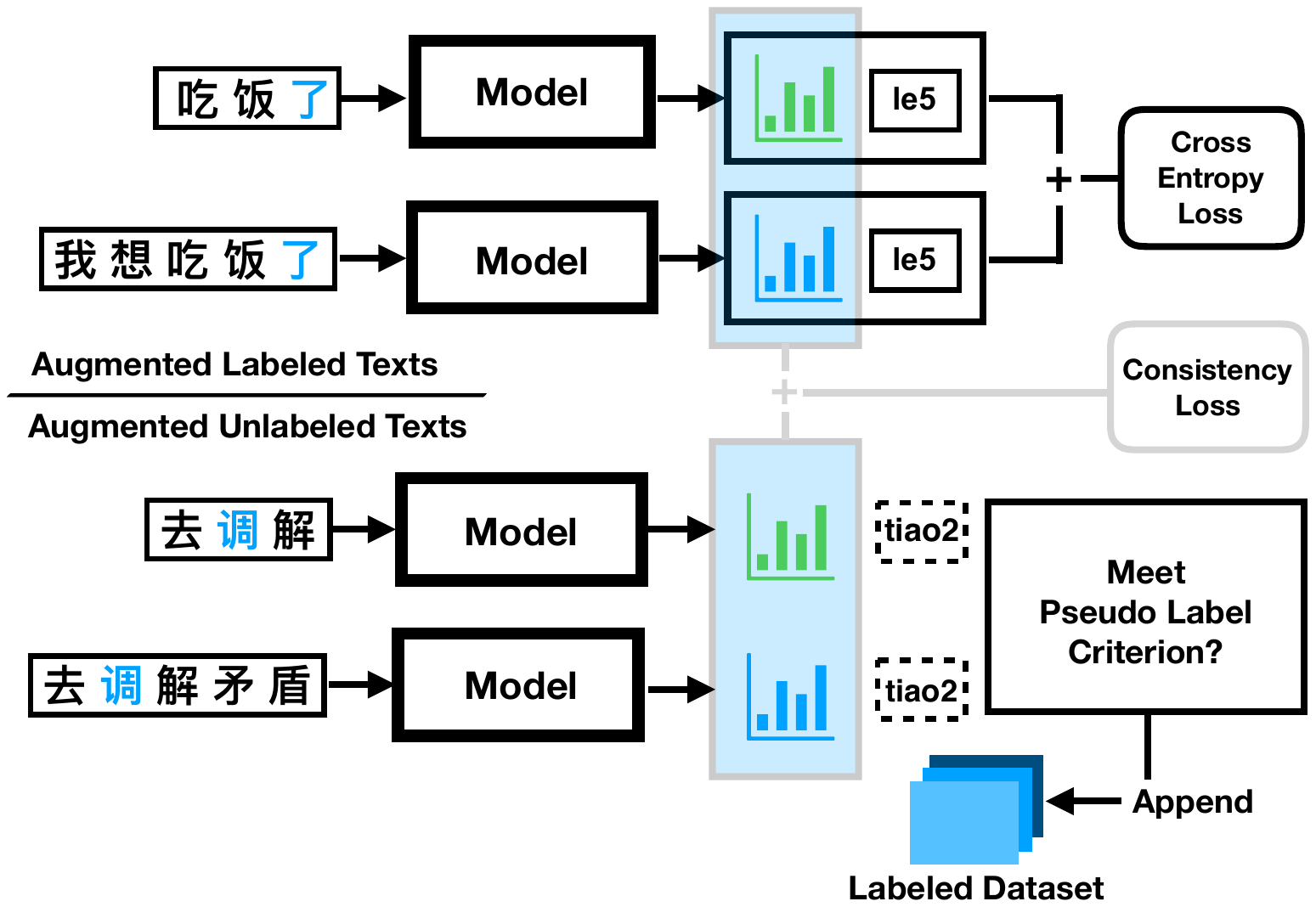}
    \caption{The framework of the semi-supervised learning on texts contained polyphonic characters. The upper part represents the training pipeline for labeled data. The lower image describes the procedure of processing unlabeled text data. }
    \label{fig:f3}
\end{figure}

\subsubsection{Consistency Regularization}\label{cl}
 Consistency Regularization is to ensure a consistent behavior when the model encounters unlabeled data similar to labeled ones.  The consistency regularization assumes that the unlabeled data and its augmented version should share a similar prediction result from a model. More specifically, input data are augmented randomly, then a loss is computed based on the comparison between the output of different versions of augmented data. The model is trained in such a self-supervised manner for generating a reliable certainty score, making the model more robust against unknown unlabeled data. 
 
The pronunciation of a polyphonic character is often irrelevant to the characters beyond a certain scope, while the influence of neighboring characters is more important. Therefore, unlike the global augmentation applied on 2D images, we ensure the text with one or several polyphonic characters keeps its key semantic information, i.e. make sure several neighbors of the target characters are unchanged. We hereby introduce three types of viable augmentation that can be applied to texts with a single target character:
\begin{enumerate}
   \item Naïve truncation: We select a subset of sequential characters in a sentence that contains the target polyphonic character. We keep a 'safe zone' of size $n$ where $n$+$n_b$ characters behind and $n$+$n_f$ in front of the target remain intact. $n_b$ and $n_f$ are random numbers ranged from zero to the maximum number of characters that existed in the corresponding direction.
    \item N-gram truncation: We first segment the sentence into words and only keep the word contained the polyphonic characters along with a random number of neighboring words. If the polyphonic character itself is a single-character word, the neighboring word is randomized between one to the maximum number of words available. Otherwise, the lower-bound of the range is set to zero.
    \item Similar words replacement: Same as the N-gram truncation strategy, we segment the sentence into words. Instead of getting discarded, each word outside the 'safe zone' has a 50\% probability to be swapped by a similar word obtained using Word2Vec.
\end{enumerate}

The above augmentation policies are executed randomly with an equal chance on any incoming text with a length larger than 3. 
In Figure \ref{fig:f3}, the consistency loss is computed simultaneously on labeled and unlabeled data. We run experiments with three different losses: Mean Square Error (MSE), Jensen–Shannon divergence (JS Divergence), and Cross Entropy. The MSE loss is calculated on the hidden vector of the last layer before the final classification block, while JS Divergence and Cross Entropy compare the probability distribution of classes.

\begin{table*}
  \caption{Benchmark of Learning Based Approaches for Polyphone Disambiguation}
  \label{tab:1}
  \centering
  \begin{tabular}{ l l l l l l l }
    \toprule
    \multicolumn{0}{c}{\textbf{Approaches}} & 
    \multicolumn{0}{c}{\textbf{Framework}} &
    \multicolumn{0}{c}{\textbf{Loss}} &
    \multicolumn{0}{c}{\textbf{Accuracy}} &
    \multicolumn{0}{c}{\textbf{Params}}  & 
    \multicolumn{0}{c}{\textbf{MAC}}  \\
    \midrule        
    
    g2pM \cite{korean} Park et al      & BLSTM                     & Cross Ent.                    & 0.749           & 421.783K      & 420.672K  \\
    MLEF \cite{MLEF} Cai et al.        & Word Embedding + BLSTM    & Cross Ent.                    & 0.843           & 1,767.104M    & 2.232M    \\
    \cite{polybert} Dai et al.         & Bert (base) + BLSTM       & Cross Ent.                    & 0.859           & 113.820M      & 113.701M  \\
    \textbf{Base}                 & Electra (tiny) + Conv + BLSTM    & Cross Ent.             & 0.864           & 15.195M       & 16.334M   \\
    \textbf{Base + C}             & Electra (tiny) + Conv + BLSTM    & Cross Ent. + Consis.   & 0.897           & -             & -         \\
    \textbf{Base + CP}            & Electra (tiny) + Conv + BLSTM    & Cross Ent. + Consis.   & \textbf{0.928}  & -             & -         \\
    
    \bottomrule
  \end{tabular}
\end{table*}

\begin{table}
  \caption{Results of Different Types of Consistency Loss}
  \label{tab:2}
  \centering
  \begin{tabular}{ l l }
    \toprule
    \multicolumn{0}{c}{\textbf{Consis. Loss Type}} & 
    \multicolumn{0}{c}{\textbf{Accuracy}}               \\
    \midrule        
    
    MSE                         & 0.926                 \\
    JS Divergence                     & 0.925                 \\
    \textbf{Cross Entropy}                & \textbf{0.928}        \\
    
    \bottomrule
  \end{tabular}
  
\end{table}

\subsubsection{Pseudo Labels}\label{pl}
Pseudo Labels are synthesized labels assigned to unlabeled data according to a procedure designed for expanding the training data. Such a labeling process is crucial since it controls the quality of assigned labels that participate in computing supervised Cross Entropy loss. In the proposed approach, we propose an entropy threshold criterion that quantifies model certainty on its prediction. In practice, pseudo labeling depends on a combined criterion that considers both confidences of model prediction and deterministic knowledge of a word-to-pinyin dictionary. 

Given the minimal cost of extracting extra knowledge from a word-to-pinyin dictionary, we label the polyphonic characters in Monophonic Character Word (MCW) with a dictionary. It serves as a procedure to speed up training and avoid erroneous labels on characters that appeared in frequent vocabulary.

Words formed by single polyphonic characters are labeled depending on the confidence of our model. In order to evaluate quantitatively the confidence of a prediction, we estimate the level of certainty based on information entropy of output probability distribution. It can be formulated as follows: 
\begin{equation}
 H(x) = -\sum_{i=0}^k P(x_i) \mathrm{log}(P(x_i))
  \label{eq3}
\end{equation}
where k stands for the number of possible pronunciations of a certain polyphonic character. Compared to setting a direct threshold on probability, calculating entropy can better reflect the confidence in our setting since different characters have a variant number of pronunciations. We set a dynamic decision boundary on entropy that gradually increase given the trust put on the model grows as the training progresses. The threshold can thus be expressed by $\mathbf{min}$(Tmax, Tmin + epoch * $\mathbf{s}$) where Tmax stands for upper bound, Tmin represents lower bound, $\mathbf{s}$ is the scale of change during each update. The update of the threshold is executed once in every two epochs.  Pseudo labels are assigned only if information entropy meets the threshold criterion.


\section{Experiment}
\subsection{Dataset}
Empirically, we find considerable unsolvable cases are among ten characters after applying a knowledge-based approach with segmentation and dictionary querying. Another problem we noticed is that the past methods perform poorly on short sentences that are frequently used in daily conversations. We prepare a dataset specifically for these ten polyphonic characters, which includes 326,000 sentences for training and a test set with 1,100 manually written and annotated texts that are deliberately made difficult for exposing model vulnerability. The number of occurrences of each character is even in both training and test sets. We plan to publish the dataset in hope of providing a thorough and convenient evaluation for future studies. As for unlabeled data, we gather 25,420,601 lines of text from Wikipedia. All texts in traditional Chinese are converted into Simplified Chinese. 


\subsection{Model Settings}
Experiments are conducted under three configurations: 1) Base model 2) Base model with consistency loss (Base model + C), 3) Base model with consistency loss while enabling pseudo labeling (Base model + CP). In configuration 2 (Base model + C), the weight of consistency loss is set as 0.008 and the weight of classification loss as 1. 

We train our base model on labeled data supervised by Cross-Entropy loss and consistency loss for 130 epochs with a batch size of 128. AdamW optimizer is applied with a learning rate of 1e-5. The output embedding size of tiny-Electra is 256. The hidden size of Conv1D and BLSTM layers is 256. We also apply a residue connection that concatenates the Electra embedding with the output of BLSTMs. 

As for configuration 3 (Base model + CP), the training setting of the base model is the same as configuration 2 (Base model + C). After we train the base model, the weight of consistency loss is modified to 0.003 and the learning rate is set to 1e-6. The consistency loss on unlabeled data is then computed and combined with consistency supervision on labeled data. We begin assigning pseudo labels to unlabeled data with the incremental entropy threshold $t$ set as Tmin = 0.81 and Tmax = 0.85, with step=0.1.  The ratio between unlabeled data and labeled data in each batch is 1:1.  We only assign pseudo labels during odd epochs, while the even epochs are for solidifying the model on data appended from the previous epochs. 

\subsection{Results and Analysis}
In this section, we compare our results with recent learning-based methods. To make a fair comparison, their model is trained on our dataset, and the weight of linear classification layers is shared among all 10 target characters. 
As depicted in Table 1, our base model outperforms all previous methods in the quantitative experiments. The proposed method even surpasses methods \cite{polybert} using base Chinese Bert with BLSTM and linear layers. The model complexity, measured by the number of parameters (Params) and Memory Access Cost (MAC), is also significantly lowered compared to the predecessor using Bert. There is a performance boost in accuracy after the introduction of consistency regularization and pseudo labeling.

\section{Conclusions}
 In this paper, we present Semi-PPL, a simple semi-supervised learning framework that leverages numerous unlabeled text data. We believe the modules mentioned in this paper are easy to be implemented in any learning-based model for Chinese character embedding learning. We make public a human-labeled test set that contains hard cases that can provide a meaningful evaluation of a Mandarin Chinese G2P system.

\bibliographystyle{IEEEtran}

\bibliography{sy.bbl}


\end{document}